\documentclass[11pt,a4paper]{article}
\usepackage[hyperref]{ranlp2023}

\usepackage{times}
\usepackage{latexsym}

\usepackage{microtype}

\usepackage{graphicx}
\usepackage{makecell}
\usepackage{amsmath}
\usepackage{amssymb}
\newcommand{\tabincell}[2]{\begin{tabular}{@{}#1@{}}#2\end{tabular}}
\aclfinalcopy

\title{Classification-Aware Neural Topic Model Combined With Interpretable
Analysis - For Conflict Classification}

\author{Tianyu Liang\textsuperscript{1}, Yida Mu \textsuperscript{1}, Soonho Kim\textsuperscript{2}, Darline Larissa Kengne Kuate\textsuperscript{2}, Julie Lang\textsuperscript{2}, Rob Vos\textsuperscript{2}, Xingyi Song\textsuperscript{1} \\
\textsuperscript{1}Department of Computer Science, University of Sheffield, Sheffield, United Kingdom \\
\textsuperscript{2}International Food Policy Research Institute, Washington, DC, USA \\
\texttt{\{tliang9, y.mu, x.song\}@sheffield.ac.uk} \\}

\begin{document}
\maketitle
\begin{abstract}
A large number of conflict events are affecting the world all the time. In order to analyse such conflict events effectively, this paper presents a Classification-Aware Neural Topic Model (CANTM-IA) for Conflict Information Classification and Topic Discovery. The model provides a reliable interpretation of classification results and discovered topics by introducing interpretability analysis. At the same time, interpretation is introduced into the model architecture to improve the classification performance of the model and to allow interpretation to focus further on the details of the data. Finally, the model architecture is optimised to reduce the complexity of the model.
\end{abstract}

\section{Introduction}

Hundreds of conflicts break out every day around the world, many of which have a major impact on the world's political and economic situation. A recent example is Ukraine Crisis, which has caused energy scarcity in Europe, a reduction in world food production and many other repercussions. For governments and institutions such as the IFPRI, the impact of conflict events can be greatly reduced if they are classified, analysed and responded to in the shortest possible time.

Our goal is to develop a deep learning model suitable for the classification of conflict information. This model should be able to classify conflict categories and discover category-related topics. Most importantly, the model must have high reliability as the consequences of conflicting information are often very serious. We therefore want to combine text classification, topic modelling and interpretable analysis to solve the problem.

Text classification assigns category labels to different texts for the purpose of distinguishing textual information. Recurrent neural networks (RNNs), convolutional neural networks (CNNs) and graph neural networks (GNNs) have all been applied to text classification tasks \citep{2015arXiv150300075T, pmlr-v37-zhub15, 2016arXiv160106733C, 2014arXiv1404.2188K, kim-2014-convolutional, johnson-zhang-2015-effective, 10.1145/3178876.3186005}. More recently, \citet{10.1007/978-3-030-32381-3_16} provides a fine-tuned BERT-based pre-training model \citep{2018BERT} for text classification tasks generic solution with new state-of-the-art results on eight extensively studied text classification datasets.

The topic model is designed to automatically find a range of topics and topic words from a collection of documents. One of the most classic topic models is latent dirichlet allocation (LDA) \citep{blei2003latent}, which is an unsupervised, non-hierarchical model. Many subsequent research has been based on LDA, such as the Hierarchical Latent Dirichlet Allocation (HLDA) proposed by \citet{NIPS2003_7b41bfa5}. In 2016, \citet{pmlr-v48-miao16} proposed a generative neural variational document model (NVDM), which models the likelihood of documents using a Variational Auto-Encoder (VAE) \citep{2013arXiv1312.6114K}. In order to purposefully uncover topic words related to the target (e.g. sentiment), many researchers have also proposed alternative approaches. For example, \citet{2018arXiv180902687D} added topic consistency to the training as part of the loss as well, thus making the latent variables dependent on the topic target as well.

Neural network-based deep learning can be described as a black box, and humans are not yet able to fully explain or peer into the entire deep learning process. So the question arises whether humans can be trusted with the decision-making mechanisms of such data-driven AI systems. The lack of interpretability leads to a reduction in the reliability of deep learning, hence the importance of interpretable analysis. In an earlier study, \citet{pmlr-v70-koh17a} hoped to find parts of the training data/training points that could be used as a basis for interpretation by introducing the influence function. Some researchers, on the other hand, have tried to find explanations for the prediction results from the test data itself. Such explanations can be found by perturbing the data \citep{2016arXiv161208220L}, extracting attention weights \citep{wiegreffe-pinter-2019-attention} or calculating the saliency scores of the input sequences \citep{2020arXiv200500115J}, etc. \citet{2016arXiv160604155L, 2020arXiv200500115J} used a combination of generators and encoders to extract rationales.

\begin{figure*}[ht]
\includegraphics[width=15cm]{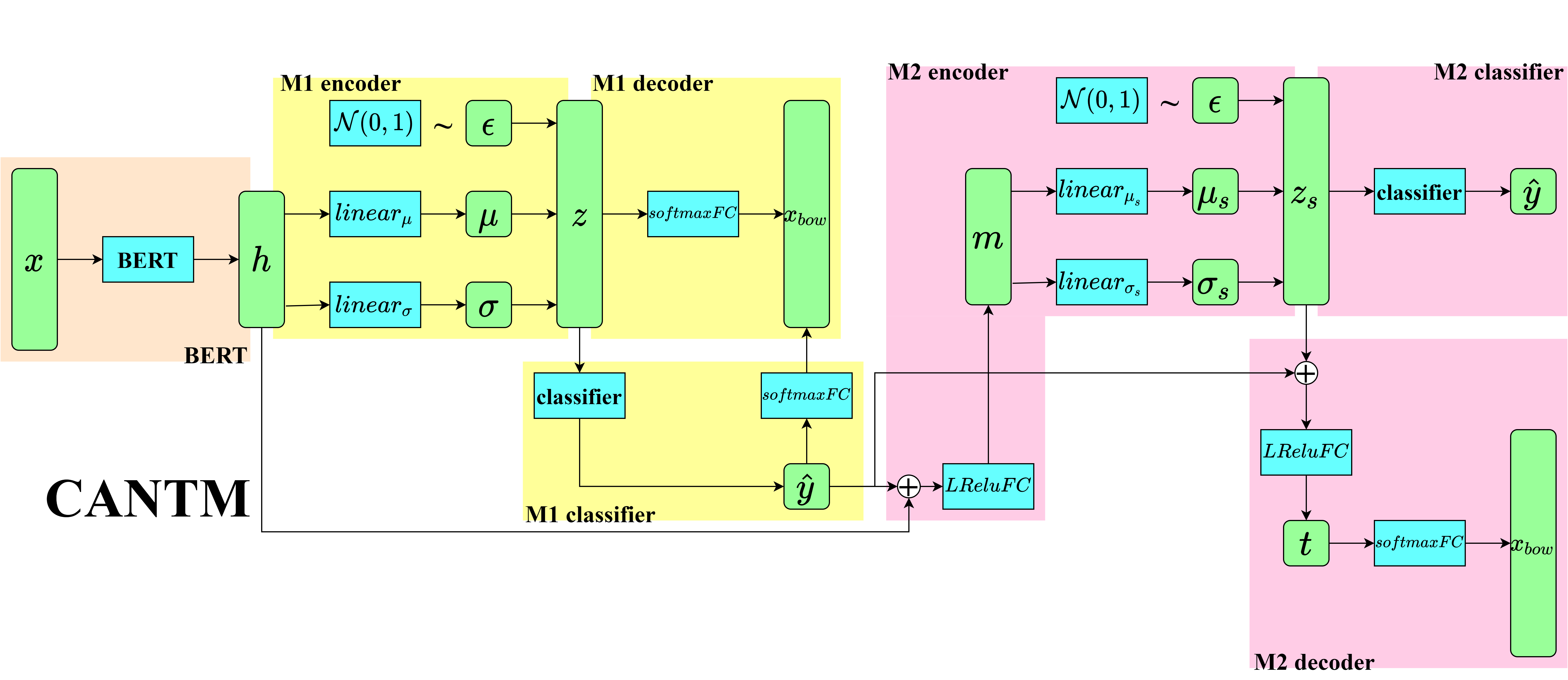}
\caption{The architecture of the CANTM.}
\label{fig:cantm}
\vspace{-15pt}
\end{figure*}

\section{Preliminary Works}

Text classification and topic modelling have been important areas of research in natural language processing. These two areas are extremely interrelated, but few studies have effectively integrated them into a unified system. One successful example is the CANTM model proposed by \citet{Reference1} on topic modelling of online text messages during the Covid-19 epidemic, which is able to effectively identify disinformation related to Covid-19 and simultaneously classify the information, helping to address issues such as citizens' distrust of government and healthcare.

The architecture of CANTM is shown in Figure \ref{fig:cantm}. The model is divided into three parts, BERT embedding, the classifier-regularised VAE (M1) and the classifier-aware VAE (M2), where the VAE architectures are used as topic models. The model first uses a BERT pre-trained model to extract segment embeddings $h$ from the input text sequence $x$. In the encoder part of M1, $h$ is transformed into the parameters $\mu$ and $\sigma$ of the Gaussian distribution via the linear layers $linear_{\mu}$ and $linear_{\sigma}$ respectively. The aim of the M1 encoder is to generate the latent variable $z$, which can be considered as hidden topics. The M1 decoder part uses the latent variable $z$ as input to reconstruct the bag of words of the input text. The M1 classifier also uses the latent variable $z$ as input, and generates classification probabilities after passing through a fully connected layer containing a softmax activation function. Note that since the classifier uses hidden topics as the basis for classification, it has not seen real data, which can reduce the overfitting of the model. The architecture of M2 is similar to M1, except that it takes the classification probabilities $\hat{y}$ output from M1 as input as well, in order to generate hidden topics $z_{s}$ guided by the classification information. Furthermore, the M2 classifier is not used to output the final classification, but only to compute joint loss during training. The joint loss function of CANTM is a combination of the loss functions of its subcomponents and is calculated as
\begin{equation}
\begin{aligned}
\mathcal{L} = \lambda\mathcal{L}_{cls} - ELBO_{x_{bow}} - ELBO_{x_{bow},\hat{y}}\\ - \mathbb{E}_{\hat{y}}[log\ p(x_{bow}|\hat{y})]
\end{aligned}
\end{equation}

CANTM has good classification and topic discovery capabilities, but it is not fully suitable for conflict information. Firstly, it does not introduce interpretability analysis to demonstrate the reliability of the model. Secondly, the topics discovered by CANTM are to some extent disturbed by a large number of neutral words present in the input text, thus making the relevance of the discovered topic words to the category information reduced. Moreover, the CANTM architecture has redundant parts, which affects its computational efficiency.

\section{Methodology}

Our model is based on an improvement of CANTM, which we call Classification-Aware Neural Topic Model Combined With Interpretable Analysis (CANTM-IA). CANTM is used as the base model because it combines text classification and topic modelling, which aligns with our goals. Secondly, the stacked VAE architecture of CANTM effectively allows us to discover the hidden topics of the target categories. In addition, topics can also be seen as an interpretation of the classification model, which facilitates our interpretability analysis and improvement of the model in conjunction with rationale.

We introduced interpretability analysis specifically by calculating the attention weights of the last layer in the BERT pre-trained model corresponding to the CLS labels and averaging them into the saliency score of the corresponding word piece. The magnitude of the saliency score is used as a visual representation of the importance of different parts of the original sample, and the parts with high saliency scores are used as the rationales of the sample. BERT parameters are frozen during training and only the last transformer encoding layer weights are unlocked for fine-tuning.

Afterwards, we use the saliency scores of the rationales instead of the bag of words of the entire input sequence as the reconstruction target in the VAE architecture. This has several advantages. First, using rationales (the part of the input sequence with high contribution) as the reconstruction target allows the topic model to focus more on the important information of the input sequence, which can reduce the interference of category-irrelevant words by the topic words and indirectly improve the classification performance of the model. Second, since the decoder uses rationales to guide the discovery of hidden topics and the classifier uses hidden topics for classification, it can be argued that these rationales explain both the hidden topics and the classification results.

In addition, there is a redundant structure in the M2 decoder part of the CANTM model. As shown in Figure \ref{fig:cantm}, $m$, as the variable that combines the input $h$ with the classification result $\hat{y}$, already introduces classification information for the rest of M2. That is, the process of generating the variable $z_{s}$ has been guided by the classification information, which generates the class-aware topics. Therefore, there is no need to reintroduce the classification result $\hat{y}$ in the decoder part of M2, and the purpose can be achieved by directly reconstructing the target using $z_{s}$ as the hidden topic variable.

\begin{figure*}[ht]
\includegraphics[width=15cm]{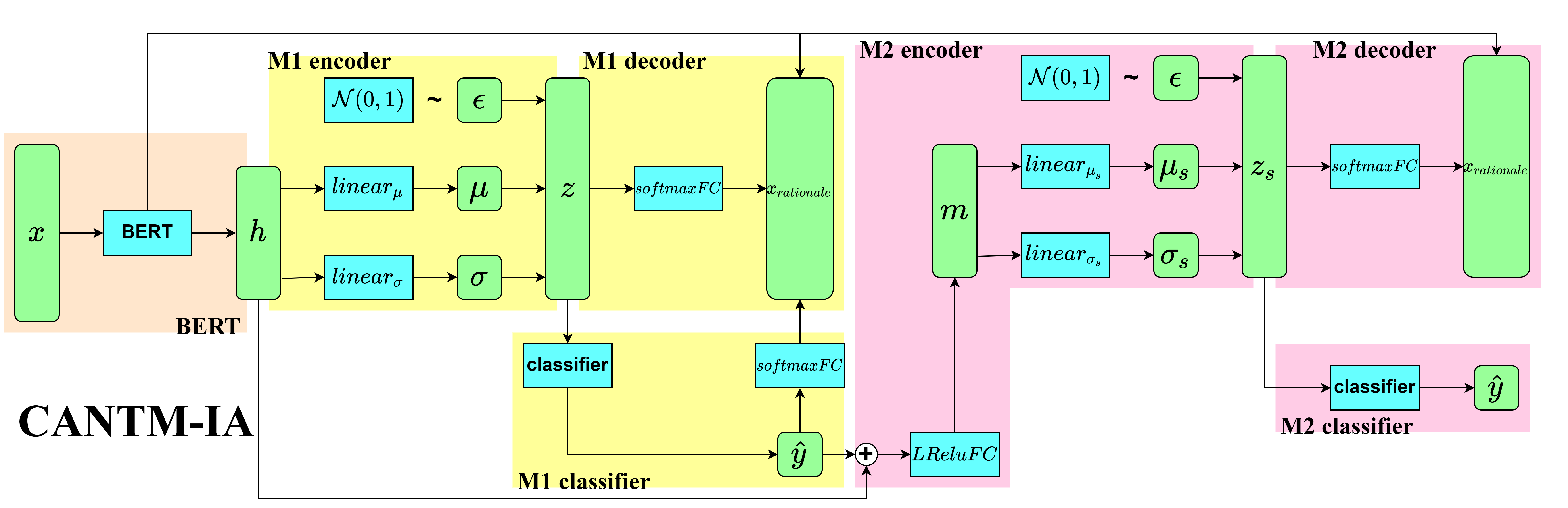}
\caption{The architecture of the CANTM-IA.}
\label{fig:cantmia}
\vspace{-5pt}
\end{figure*}

Combining the above optimisation methods, the modified CANTM-IA model is shown in Figure \ref{fig:cantmia}.

\section{Experiments}

\subsection{Dataset}

We use The Armed Conflict Location \& Event Data Project (ACLED), a disaggregated data collection, analysis, and crisis mapping project, as our source dataset \cite{doi:10.1177/0022343310378914}. The ACLED dataset collects six types of events. We use data spanning a full 3 years between 25 June 2019 and 24 June 2022 as experimental data. Of these, the volume of data for the conflict category Protests is 415,588, which far exceeds the volume of data for the other categories. In order to ensure a balanced dataset, a quarter of the data, i.e. 103,897 items, are randomly selected as the data of category Protests for the experiment. In addition, 50,000 texts from WMT News Crawl Dataset \footnote{\url{http://www.statmt.org/wmt13/training-monolingual-news-2012.tgz}} are used as the out-of-domain data. The details of the experimental dataset are shown in table \ref{tab:dataset}, with 90.43\% of the ACLED data and 9.57\% of the regular news data. The training set, validation set and test set are sampled from the original dataset in a 7:1:2 ratio

\begin{table*}
\centering
\small
\begin{tabular}{ccccccccc}
\hline
\tabincell{c}{Type of\\conflict} & \tabincell{c}{Battles} & \tabincell{c}{Explosions/\\Remote violence} & \tabincell{c}{Protests} & \tabincell{c}{Riots} & \tabincell{c}{Strategic\\developments} & \tabincell{c}{Violence agai-\\nst civilians} & \tabincell{c}{Out of\\domain} & \tabincell{c}{Total}\\
\hline
Train & 76202 & 61222 & 72727 & 34097 & 30294 & 56329 & 35000 & 365871\\
Valid & 10887 & 8747 & 10390 & 4872 & 4328 & 8047 & 5000 & 52271\\
Test & 21772 & 17493 & 20780 & 9744 & 8656 & 16094 & 10000 & 104539\\
\hline
Total & 108861 & 87462 & 103897 & 48713 & 43278 & 80470 & 50000 & 522681\\
\hline
\end{tabular}
\caption{Information of the experimental data set.}
\label{tab:dataset}
\vspace{-15pt}
\end{table*}

\subsection{Experimental Setup}

We compare our CANTM-IA model with two strong baseline models: {\bf BERT} and {\bf CANTM}. For BERT model, a linear layer of dimension 300 is connected to BERT [CLS] Token output and uses a fully-connected layer with a softmax activation function as a classifier to output the classification results. For CANTM model, we using a bag of words of size 500 and a hidden topic variable of dimension 100.     

Three sets of experiments are conducted to compare the choice of parameters and the impact on {\bf CANTM-IA}. The first set uses rationales with a ratio of 10\% of the number of tokens in the input text as the reconstruction target, denoted as CANTM-IA (ratio 0.1). In the second set, this proportion is 50\% and is denoted as CANTM-IA (ratio 0.5). In addition, a fine-tuning experiment is carried out to fine-tune the model parameters using the CANTM-IA architecture on the trained CANTM model for only 1 epoch. The rationales used for the fine-tuning experiment are scaled to 50\% and the model is denoted as CANTM-IA (fine-tune). Other model parameters are kept consistent with the CANTM baseline system.

We use BERT-base-uncased in experiments, only the last transformer encoding layer is unlocked for fine-tuning, and remaining BERT parameters are frozen during training.

\subsection{Results}

The overall classification results are shown in Table \ref{tab:classification_result}. BERT is a strong baseline with a solid classification accuracy (0.9738). On this basis, CANTM and CANTM-IA still obtained better classification performance by using hidden topics as the basis for classification. The best performing CANTM-IA (ratio 0.5) model achieved an accuracy of 0.9780 and an F1 score of 0.9791, which demonstrates the effectiveness of using hidden topics as a basis for classification. Furthermore, the classification performance of the CANTM-IA (fine-tune) is improved over the CANTM model, even after only 1 fine-tuning. This suggests a positive contribution of the topic model guided by rationale to the effectiveness of text classification. The F1 scores for each sub-category in the dataset are given in Table \ref{tab:classification_result_f1_each_class}.

\begin{table}
\centering
\small
\begin{tabular}{ccc}
\hline
\tabincell{c}{Model} & \tabincell{c}{Accuracy} & \tabincell{c}{F-1}\\
\hline
BERT (baseline) & 0.9738 & 0.9749\\
CANTM (baseline) & 0.9751 & 0.9760\\
CANTM-IA (fine-tune) & 0.9766 & 0.9775\\
CANTM-IA (ratio 0.1) & 0.9774 & 0.9787\\
CANTM-IA (ratio 0.5) & \textbf{0.9780} & \textbf{0.9791}\\
\hline
\end{tabular}
\caption{Comparison of the classification performance.}
\label{tab:classification_result}
\vspace{-15pt}
\end{table}

\begin{table*}
\centering
\small
\begin{tabular}{cccccccc}
\hline
\tabincell{c}{Model} & \tabincell{c}{Battles} & \tabincell{c}{Explosions/\\Remote violence} & \tabincell{c}{Protests} & \tabincell{c}{Riots} & \tabincell{c}{Strategic\\developments} & \tabincell{c}{Violence agai-\\nst civilians} & \tabincell{c}{Out of\\domain}\\
\hline
\tabincell{c}{BERT (baseline)} & 0.9583 & 0.9817 & 0.9904 & 0.9754 & 0.9693 & 0.9501 & 0.9994\\
\tabincell{c}{CANTM (baseline)} & 0.9628 & 0.9836 & 0.9879 & 0.9707 & 0.9736 & 0.9540 & \textbf{0.9998}\\
\tabincell{c}{CANTM-IA (fine-tune)} & 0.9646 & \textbf{0.9854} & 0.9885 & 0.9704 & 0.9769 & 0.9575 & 0.9996\\
\tabincell{c}{CANTM-IA (ratio 0.1)} & 0.9633 & 0.9849 & 0.9910 & 0.9769 & \textbf{0.9777} & 0.9570 & 0.9997\\
\tabincell{c}{CANTM-IA (ratio 0.5)} & \textbf{0.9655} & 0.9847 & \textbf{0.9911} & \textbf{0.9772} & 0.9774 & \textbf{0.9584} & 0.9995\\
\hline
\end{tabular}
\vspace{-5pt}
\caption{F1 scores of the models for different categories of classification results.}
\label{tab:classification_result_f1_each_class}
\vspace{-5pt}
\end{table*}

It should be noted that since the ACLED data is cleaned by a professional data agency, the content of the data is to a large extent highly normative and accurate. As a result, classification performance can be extremely good even for the baseline model. This makes it appear that the improved model cannot outperform the baseline model by much in terms of experimental results. However, in this case, due to the large amount of data in the dataset, even a subtle advantage is evident in the face of the number of accurate predictions.

\begin{table*}
\centering
\small
\begin{tabular}{cll}
\hline
\makecell[c]{Type of conflict} & \makecell[c]{Topic words in CANTM} & \makecell[c]{Topic words in CANTM-IA (ratio 0.5)}\\
\hline
\tabincell{c}{Battles} & \tabincell{l}{forces military fatalities killed positions\\clashed militants taliban coded azerbaijan} & \tabincell{l}{clashed killed clashes fire attacked\\clash fired attack small militants}\\
\hline
\tabincell{c}{Explosions/\\Remote violence} & \tabincell{l}{shelled forces fatalities injuries  unknown\\positions artillery smm osce airstrikes} & \tabincell{l}{shelled casualties fired targeted fatalities\\total airstrikes artillery killed carried}\\
\hline
\tabincell{c}{Protests} & \tabincell{l}{report protest people city government\\members held workers gathered protested} & \tabincell{l}{protest protested demonstrated held protesters\\gathered staged demonstration workers demanding}\\
\hline
\tabincell{c}{Riots} & \tabincell{l}{report police rioters demonstrators demon-\\stration clashed group people stones injured} & \tabincell{l}{rioters demonstrators clashed demonstration\\set attacked beaten beat burning fire}\\
\hline
\tabincell{c}{Strategic\\developments} & \tabincell{l}{property destruction forces military arrested\\township district seized movement security} & \tabincell{l}{arrested set destroyed looted fire seized\\military destruction burned forces}\\
\hline
\tabincell{c}{Violence\\against civilians} & \tabincell{l}{killed shot man men fatality armed\\ found colonia unidentified body} & \tabincell{l}{killed shot man attacked armed\\people found beat abducted dead}\\
\hline
\end{tabular}
\vspace{-5pt}
\caption{Top 10 topic words for each conflict type.}
\label{tab:cantm_topic}
\vspace{-15pt}
\end{table*}

We show the top 10 topic words for each conflict category for the CANTM and CANTM-IA (ratio 0.5) models in table \ref{tab:cantm_topic}. As can be seen, the category-related topic words extracted by CANTM already provide a good overview for each conflict category. However, there are still many neutral words such as 'report', 'city' and 'unknown' in the CANTM results. This is due to the fact that CANTM uses a bag of words from the complete input sequence for topic reconstruction, which makes a large number of neutral words that appear in the conflict text influential in the reconstruction process. The weights of the reconstruction matrix with respect to this token is strengthened during training, and the relevance of this word to the relevant category is increased. In contrast, CANTM-IA cleverly reduces the influence of such neutral words. Because CANTM-IA uses rationales as the reconstruction target, this allows the model to focus more on the conflict-related information itself and thus ignore irrelevant neutral words. This results in a greater concentration of topic words that are relevant to the classification results and more representative of the categories. It also demonstrates the effectiveness of category-related topic words extraction and ensures the possibility of subsequent analysis of the model.

\begin{table}
\centering
\small
\begin{tabular}[7.5cm]{cl}
\hline
\makecell[c]{Model} & \makecell[c]{Note of event (with highlighted rationales)}\\
\hline
\tabincell{c}{BERT\\(baseline)} & \tabincell{l}{{\color{red}Property destruction:} Around 13 May 2022\\(as reported), in Ta Tang Ku village of Pin-\\laung township (coded as Pinlaung) (Pa-O\\Self-Administered {\color{red}Zone}, Shan-South state),\\the PNO {\color{red}soldiers} {\color{red}destroyed} the Catholic\\statue of the Virgin Mary and a Catholic\\church. The statue is well {\color{red}respected}.}\\
\hline
\tabincell{c}{CANTM\\(baseline)} & \tabincell{l}{Property destruction: Around 13 May 2022\\({\color{red}as} {\color{red}reported}), in Ta Tang Ku village {\color{red}of} Pin-\\laung township (coded as Pinlaung) (Pa-O\\Self-Administered {\color{red}Zone}, Shan-South state),\\the PNO {\color{red}soldiers} {\color{red}destroyed} the Catholic\\statue of the Virgin Mary and a Catholic\\{\color{red}church}. The statue is well respected.}\\
\hline
\tabincell{c}{CANTM-IA\\(ratio 0.5)} & \tabincell{l}{Property destruction: Around 13 May 2022\\(as reported), in Ta Tang Ku village of Pin-\\laung township (coded as Pinlaung) (Pa-O\\Self-Administered Zone, Shan-South state),\\{\color{red}the} PNO {\color{red}soldiers} {\color{red}destroyed} the Catholic\\statue of the Virgin Mary and {\color{red}a} Catholic\\{\color{red}church}. The statue is well respected.}\\
\hline
\end{tabular}
\vspace{-5pt}
\caption{Comparison of rationales extracted from the same sample. Words in red are rationales.}
\label{tab:rational}
\vspace{-15pt}
\end{table}

We show rationale examples in table \ref{tab:rational} that are extracted by the different models from the same sample. It can be observed that while the rationales extracted by CANTM from can focus on conflict-related information ("soldiers", "destroyed", "church"), there is also some irrelevant information that is focused on ("as", "reported", "of", "zone"). CANTM-IA (ratio 0.5), on the other hand, focuses precisely and intently on the conflict information itself ("the", "soldiers", "destroyed", "a", "church"). Note that although the words "the" and "a" appear to be meaningless and category-independent words on their own. However, the model incorporates contextual information. Therefore, it can be argued that the CANTM-IA model also combines and pays some attention to coherent semantics, which makes the rationales extracted by CANTM-IA more coherent than previous rationales, and allows for better interpretation of the model's classification decisions and topic selection. The rationales comparison experiment shows that the rationales extracted by CANTM-IA focuses on the conflict information itself and can reasonably and effectively explain the model's conflict type classification results. This ensures the reliability of the model's classification decisions and allows CANTM-IA to provide humans with reliable results for further analysis of conflict information to a certain extent.

\section{Conclusion}

We proposed a Classification-Aware Neural Topic Model (CANTM-IA) for Conflict Information Classification and Topic Discovery in this paper. The classification results and topic models of CANTM-IA can be reliably interpreted using rationales. Also, rationales are introduced into the topic model to improve model performance. Finally, the model architecture has been optimised. Compared to the baseline systems, CANTM-IA has improved predictive performance, reliability and efficiency. Our future work will be to adapt the model to other types of data and to refine the way in which interpretable analysis is introduced.

\section{Ethics and Broader Impact Statement}
\subsection{Ethics}
Only publicly available dataset\footnote{\url{https://www.prio.org/misc/Download.aspx?file=\%2fcscw\%2frd\%2fReplication+Data\%2fReplication+data_Raleigh+et+al+47(5).zip}} is used in this paper \citet{doi:10.1177/0022343310378914}. No ethical approval is required for this work.
\subsection{Implications}
Our work has several potential practical implications:
\begin{itemize}
    \item Our model outperforms two strong baselines, BERT and CANTM, in terms of predictive performance. It can also serve as a competitive baseline for future research.
    \item Our explainable neural topic model, CANTM-IA, can be utilized for other NLP downstream tasks, such as stance detection \citep{mu2023vaxxhesitancy} and rumor verification \citep{derczynski-etal-2017-semeval}, providing \textbf{interpretable predictions}.
\end{itemize}

\section*{Acknowledgments}

This research is partially supported by an European Union Horizon 2020 Project (Agreement no.871042 under the scheme “INFRAIA-01-2018-2019 – Integrating Activities for Advanced Communities”: “SoBigData++: European Integrated Infrastructure for Social Mining and Big Data Analytics” (http://www.sobigdata.eu)).

\bibliographystyle{acl_natbib}
\bibliography{custom}
\end{document}